\documentclass[conference]{IEEEtran}

\usepackage{cite}
\usepackage{amsmath,amssymb,amsfonts}
\usepackage{graphicx}
\usepackage{textcomp}
\usepackage{xcolor}
\usepackage{booktabs}
\usepackage{multirow}
\usepackage{array}
\usepackage{tabularx}
\usepackage{url}
\usepackage{subcaption}
\usepackage{adjustbox}
\usepackage{float}
\usepackage{longtable}
\usepackage[table]{xcolor}
\usepackage[dvipsnames]{xcolor}

\begin{document}

\title{Semantic Context Matters: Analysis of Color Names Across Domains}

\author{
\IEEEauthorblockN{Author Name}
\IEEEauthorblockA{
Department / University / Organization\\
City, Country\\
email@example.com}
}

\author{
\IEEEauthorblockN{Adilet Yerkin, Elnara Kadyrgali, Malika Ziyada, Nuray Toganas, \\ Muragul Muratbekova,  Ayan Igali, Aruzhan Sabitkyzy, Pakizar Shamoi\IEEEauthorrefmark{1}}
\IEEEauthorblockA{School of Information Technology and Engineering,\\
Kazakh-British Technical University,
Almaty, Kazakhstan 050000\\ 
Email: p.shamoi@kbtu.kz
}
}

\maketitle

\begin{abstract}
Color naming is influenced not only by physical color values but also by the semantic context in which colors are used. This paper investigates context-dependent color naming by mapping color-name datasets from Cosmetics, Crayola, and Car-color vocabularies onto the 86 fuzzy color categories of the COLIBRI color model.  Contextual variation is analyzed using category coverage, Shannon entropy, and maximum lift. The results show that the three contexts occupy the COLIBRI color space differently: Cosmetics covers 48 of 86 fuzzy categories, Crayola covers 50, and Car colors cover 40. The results demonstrated that Crayola provides the broadest and most balanced use of the fuzzy color space, Cosmetics is mainly concentrated around warm-tone regions, and Car colors are more specialized around blue and achromatic regions. These findings show that color naming cannot be fully explained by numerical color similarity alone and that semantic context plays an important role in human color interpretation. The proposed framework supports the development of context-aware color models for design analytics, product search, recommendation systems, and human-centered artificial intelligence.



\end{abstract}


\section{Introduction}
Color is one of the most important perceptual attributes used in visual communication, marketing, digital design, search systems, and recommendation platforms \cite{OConnor2015Colour, Sakib2026The}. In computational systems, colors are usually represented through numerical spaces such as RGB, HSV, HSI, and CIELAB, which allow colors to be compared using mathematical distance or similarity measures.

However, numerical similarity does not always correspond to semantic similarity \cite{Bodrogi2014Semantic}. One color may receive different names depending on the object domain, usage context, contrast, or cultural and commercial vocabulary\cite{Meliksetyan2025THE, Sakib2026The}. So, mathematically identical or highly similar colors may receive different annotations depending on object context

Different domains develop their own color vocabularies \cite{gibson2017coloruse, Twomey2021What}. For example, cosmetics use shade names associated with appearance, skin tone, mood, and marketing appeal, such as nude, rose, coral, or wine. Children’s art materials often use expressive and educational color names, while car catalogs frequently emphasize neutral, metallic, gray, blue, and dark colors. Therefore, each domain may occupy and partition the perceptual color space differently.

Previous studies have investigated color naming, basic color terms, and perceptual color categorization. However, less attention has been given to quantitatively comparing how different semantic color-name vocabularies occupy the same fuzzy perceptual color space.

To address this gap, this study maps color-name datasets from \textit{Crayola},  \textit{Car} and \textit{Cosmetics}-related vocabularies into the 86 fuzzy color categories of the COLIBRI model \cite{shamoi2025colibri}. Each context is represented as a normalized 86-dimensional fuzzy-color distribution. 

The main contributions of this paper are as follows:
\begin{itemize}
    \item We propose a framework for comparing domain-specific color-name vocabularies in a shared fuzzy perceptual color space
    \item We introduce a rule-based color naming richness classification scheme based on coverage, Shannon entropy, and maximum lift, enabling comparison of color vocabularies across semantic domains.
    \item 
    We conduct a case study using datasets from \textit{Cosmetics}, an art supplier (\textit{Crayola}), and \textit{Car} color to investigate whether semantic context affects color naming. 
\end{itemize}

The remainder of this paper is organized as follows. Section II reviews related work on color naming and context-dependent color interpretation. Section III describes the datasets used in this study and presents the proposed methodology. 
Section IV reports and discusses the experimental results across the four semantic contexts. Finally, Section V concludes the paper and outlines directions for future work.

\section{Related Work}

\begin{figure*}[ht!]
    \centering
    \includegraphics[width=\textwidth]{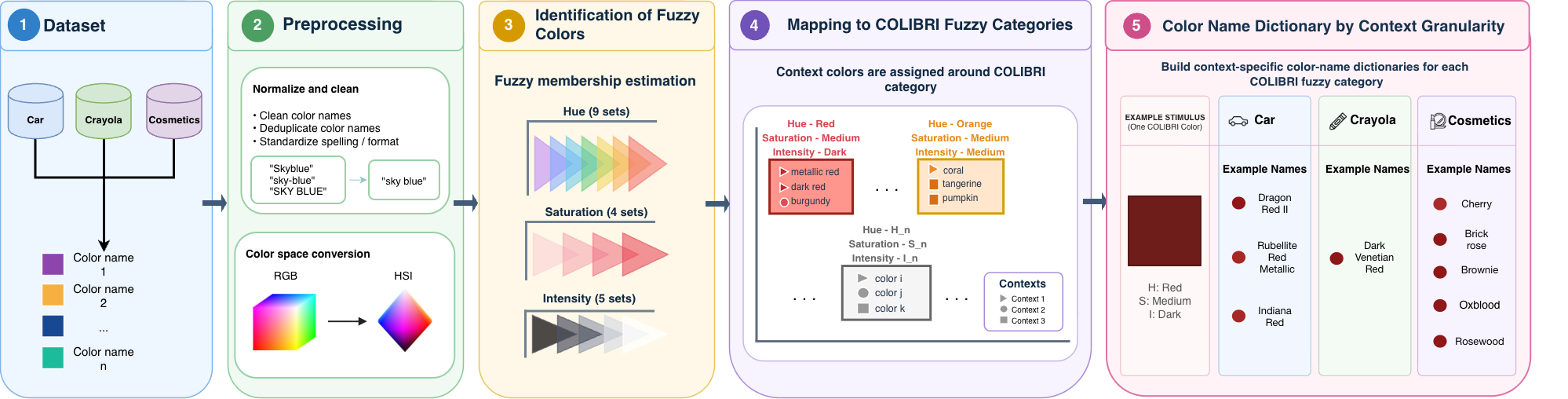}
    \caption{The pipeline preserves the fixed COLIBRI perceptual structure while allowing domain-specific names and samples to be placed as soft members of one or more fuzzy color categories.}
    \label{fig:methodology_figure}
\end{figure*}

Color naming plays a fundamental role in visual communication, enabling people to describe, categorize, and interpret colors in a meaningful way. Berlin and Kay \cite{berlin1991basic} explored basic color terms across languages and identified universal semantics. Their study found that human visual physiology impacts the evolution of color vocabularies, resulting in a limited number of composite color categories. Another study addressed domain-specific color naming and argued that many applications require color vocabularies beyond the 11 basic color terms \cite{yu2018weaklysuperviseddomainspecificcolor}; their weakly supervised, attention-based model learned color names across domains from weak image labels.


Multiple studies show that semantically meaningful colors improve the interpretation of categorical visualizations \cite{lin2013selecting}, \cite{schloss2020semantic}, \cite{mukherjee2021context}. Building on this, subsequent research has investigated how colors are perceived and categorized in different visual contexts. Studies on color appearance descriptors identified categories such as Bright, Vivid, Strong, Dull, Pale, and Dark, with Bright consistently recognized more reliably than Dull and Pale across different experiments \cite{giragama2006cross,othman2020categorizing}.


Beyond color perception and categorization, Lindsey and Brown \cite{lindsey2009wcs} demonstrated that color naming follows a limited set of recurring motifs across languages. Their study revealed well-defined regions of the color space where speakers agree strongly on a name, separated by boundary regions of low naming agreement. This unevenness is not arbitrary: warm colors such as reds and yellows are consistently communicated more efficiently than cool colors, an asymmetry that mirrors the color statistics of salient real-world objects \cite{gibson2017coloruse}. Related work formalizes this using information theory, showing that color-naming systems naturally exhibit soft category boundaries, often leaving parts of the color space without a single dominant name \cite{zaslavsky2018efficient}. Next, the COLIBRI model represents colors using fuzzy linguistic categories, providing a more human-centered representation of color perception \cite{shamoi2025colibri}. Color meaning is flexible in communication, category judgments are most stable at category centers and less stable at boundaries, and perception shifts with context ranging from displays and visualizations to retail settings and local visual environments \cite{schloss2020semantic}. Taken together, these findings indicate that the consistency with which a color is named depends on how that color is used, and therefore that the distribution of names across a fixed set of colors may differ from one usage context to another.

Color naming also takes concrete form in applied, domain-specific naming systems, e.g., the standardized set of named colors used on the web, where a fixed vocabulary of color keywords is defined for describing colors \cite{csscolor4}. Also, the next study proposes a framework for assigning semantically meaningful colors to category names using linguistic information, semantic analysis, and image retrieval \cite{setlur2015linguistic}. Another study examined the role of surface color, brightness, and texture gradients in object categorization and naming, showing that surface information contributes to how objects are identified and described \cite{Price1989}.

Other domains have developed their own naming conventions, from artists' and manufacturers' color ranges to industry catalogs of product colors. Each such system is a self-contained vocabulary shaped by the needs of its own domain, and different naming systems partition and cover the color space in different ways. This makes curated, domain-specific naming systems a natural basis for examining how color naming varies across contexts.

Overall, previous studies demonstrate that color naming is influenced by perceptual and linguistic factors. These findings provide a foundation for exploring the role of the context in color interpretation.








\section{Methodology}

\subsection{Dataset}


The dataset was collected across three domains: art suppliers, cars, and cosmetics.  These domains are selected because they differ in visual semantics, material characteristics, and typical color vocabulary. 

For the art supplier domain, data were obtained via web scraping of publicly available color reference sources (\textit{Crayola}), extracting RGB values along with their corresponding color names. For the cosmetics domain, shade names and color information were collected from the cosmetics' website, a major online beauty products retailer, across product categories including lipsticks, concealers, eyeliners, mascaras, and blushes.

During collection, shade names that were primarily evocative or marketing-driven rather than descriptive of the actual color (e.g., names referencing emotions, moods, or abstract concepts with no direct color correspondence) were excluded, and only names that genuinely describe the perceived color were retained. For the car domain, exterior paint colors were collected for eight manufacturers -- Tesla, BMW, Audi, Zeekr, Volkswagen, Mercedes-Benz, Kia, and Bentley -- using each manufacturer's officially published color names and corresponding RGB values. Table~\ref{tab:initial_dataset_example} presents a sample of the resulting dataset structure prior to preprocessing.


\begin{table}[!t]
\centering
\caption{Example of the initial dataset structure before preprocessing.}
\label{tab:initial_dataset_example}
\footnotesize
\renewcommand{\arraystretch}{1.10}
\setlength{\tabcolsep}{4pt}
\begin{tabular}{lll}
\toprule
\textbf{Context} & \textbf{RGB value} & \textbf{Name of color} \\
\midrule
Cosmetics & \texttt{[219, 157, 150]} & pale pink \\
Crayola   & \texttt{[236, 235, 189]} & spring green \\
Car       & \texttt{[28, 79, 115]}   & sea blue metallic \\
\ldots    & \ldots                   & \ldots \\
\bottomrule
\end{tabular}
\end{table}







\subsection{Proposed approach}

\begin{table}[!t]
\centering
\caption{Representative examples of color names mapped to selected fuzzy color regions across semantic contexts.}
\label{tab:selected_fuzzy_examples}
\scriptsize
\renewcommand{\arraystretch}{1.08}
\setlength{\tabcolsep}{1.2pt}
\begin{tabularx}{\columnwidth}{>{\centering\arraybackslash}p{0.045\columnwidth} >{\centering\arraybackslash}p{0.065\columnwidth} >{\raggedright\arraybackslash}p{0.175\columnwidth} >{\raggedright\arraybackslash}X >{\raggedright\arraybackslash}X >{\raggedright\arraybackslash}X}
\toprule
\textbf{ID} & \textbf{Color} & \textbf{Fuzzy name (H S I)} & \textbf{Cosmetics} & \textbf{Crayola} & \textbf{Car} \\
\midrule
2  & \fcolorbox[HTML]{686868}{686868}{\rule{0pt}{0.75em}\rule{1.1em}{0pt}} & dark gray & pitch black; black metal & black & Phantom Black; Deep Black Pearlescent \\
8  & \fcolorbox[HTML]{FF7777}{FF7777}{\rule{0pt}{0.75em}\rule{1.1em}{0pt}} & red medium dark & Oxblood; Bronze & dark venetian red & Indiana Red Metallic; Copper Metallic \\
10 & \fcolorbox[HTML]{FF0E0E}{FF0E0E}{\rule{0pt}{0.75em}\rule{1.1em}{0pt}} & red medium light & Red; Cotton mauve & bittersweet; vivid tangerine & -- \\
15 & \fcolorbox[HTML]{76432C}{76432C}{\rule{0pt}{0.75em}\rule{1.1em}{0pt}} & orange low medium & Brown sugar; Latte & beaver; raw umber & Merian Brown; Dakota Beige \\
18 & \fcolorbox[HTML]{A23A0A}{A23A0A}{\rule{0pt}{0.75em}\rule{1.1em}{0pt}} & orange medium medium & Espresso; Nude truffle & burnt sienna; brown & Orange Flame Satin by Mulliner \\
28 & \fcolorbox[HTML]{B48A0E}{B48A0E}{\rule{0pt}{0.75em}\rule{1.1em}{0pt}} & yellow medium light & Mellow yellow & maize; orange-yellow & -- \\
34 & \fcolorbox[HTML]{67A440}{67A440}{\rule{0pt}{0.75em}\rule{1.1em}{0pt}} & green low light & green; Tropic Green & spring green; granny smith apple & Light Emerald Legacy; Patina \\
36 & \fcolorbox[HTML]{429A0A}{429A0A}{\rule{0pt}{0.75em}\rule{1.1em}{0pt}} & green medium medium & emerald cut; Forest green & olive green; maximum green & -- \\
55 & \fcolorbox[HTML]{0E96A8}{0E96A8}{\rule{0pt}{0.75em}\rule{1.1em}{0pt}} & light blue medium light & blue trip; Blue & turquoise blue; sky blue & -- \\
63 & \fcolorbox[HTML]{0A30AC}{0A30AC}{\rule{0pt}{0.75em}\rule{1.1em}{0pt}} & blue medium medium & Maya blue; Ultramarine & blue; bluetiful & Frost Blue Metallic; Neptune \\
70 & \fcolorbox[HTML]{72409A}{72409A}{\rule{0pt}{0.75em}\rule{1.1em}{0pt}} & violet low light & Lavender haze & wisteria; lavender & -- \\

\bottomrule
\end{tabularx}
\end{table}


\subsubsection{Procedure}

Each RGB color sample from the dataset is first transformed into the HSI representation and then mapped to the corresponding color category in the COLIBRI model, which consists of 86 predefined fuzzy color classes \cite{shamoi2025colibri}. The original textual color names provided in the datasets are preserved and associated with the assigned COLIBRI categories for subsequent analysis, as shown in Table \ref{tab:selected_fuzzy_examples}.

The overall processing pipeline is summarized as
\begin{equation}
\mathrm{RGB} \rightarrow \mathrm{HSI} \rightarrow \mathrm{COLIBRI}.
\end{equation}







After the mapping stage, all textual color names assigned to the same COLIBRI color category are grouped separately for each application domain. 




\begin{table*}[ht!]
\centering
\caption{Coverage, entropy, lift, and richness classification across semantic color contexts.}
\label{tab:coverage_entropy_metrics}
\begin{adjustbox}{width=\textwidth}
\begin{tabular}{lrrrrrrrrl}
\toprule
\textbf{Context} &
\textbf{\begin{tabular}[c]{@{}c@{}}Total\\ Count\end{tabular}} &
\textbf{\begin{tabular}[c]{@{}c@{}}Represented\\ Fuzzy Colors\\ out of 86\end{tabular}} &
\textbf{\begin{tabular}[c]{@{}c@{}}Coverage\\ out of 86\\ (\%)\end{tabular}} &
\textbf{\begin{tabular}[c]{@{}c@{}}Shannon\\ Entropy\end{tabular}} &
\textbf{\begin{tabular}[c]{@{}c@{}}Normalized\\ Entropy\end{tabular}} &
\textbf{\begin{tabular}[c]{@{}c@{}}Effective\\ Number of\\ Fuzzy Colors\end{tabular}} &
\textbf{\begin{tabular}[c]{@{}c@{}}Max\\ Probability\end{tabular}} &
\textbf{\begin{tabular}[c]{@{}c@{}}Max\\ Lift\end{tabular}} &
\textbf{\begin{tabular}[c]{@{}c@{}}Richness\\ Class\end{tabular}} \\
\midrule
Cosmetics & 176 & 48 & 55.81 & 4.75 & 0.74 & 26.83 & 0.16 & 14.17 & Broad-concentrated \\
Crayola   & 169 & 50 & 58.14 & 5.31 & 0.83 & 39.66 & 0.07 & 6.11  & Broad-balanced \\
Car       & 297 & 40 & 46.51 & 4.48 & 0.70 & 22.29 & 0.14 & 12.45 & Narrow-specialized \\
\bottomrule
\end{tabular}
\end{adjustbox}
\end{table*}

\begin{figure}[tb]
    \centering
    \includegraphics[width=0.9\columnwidth]{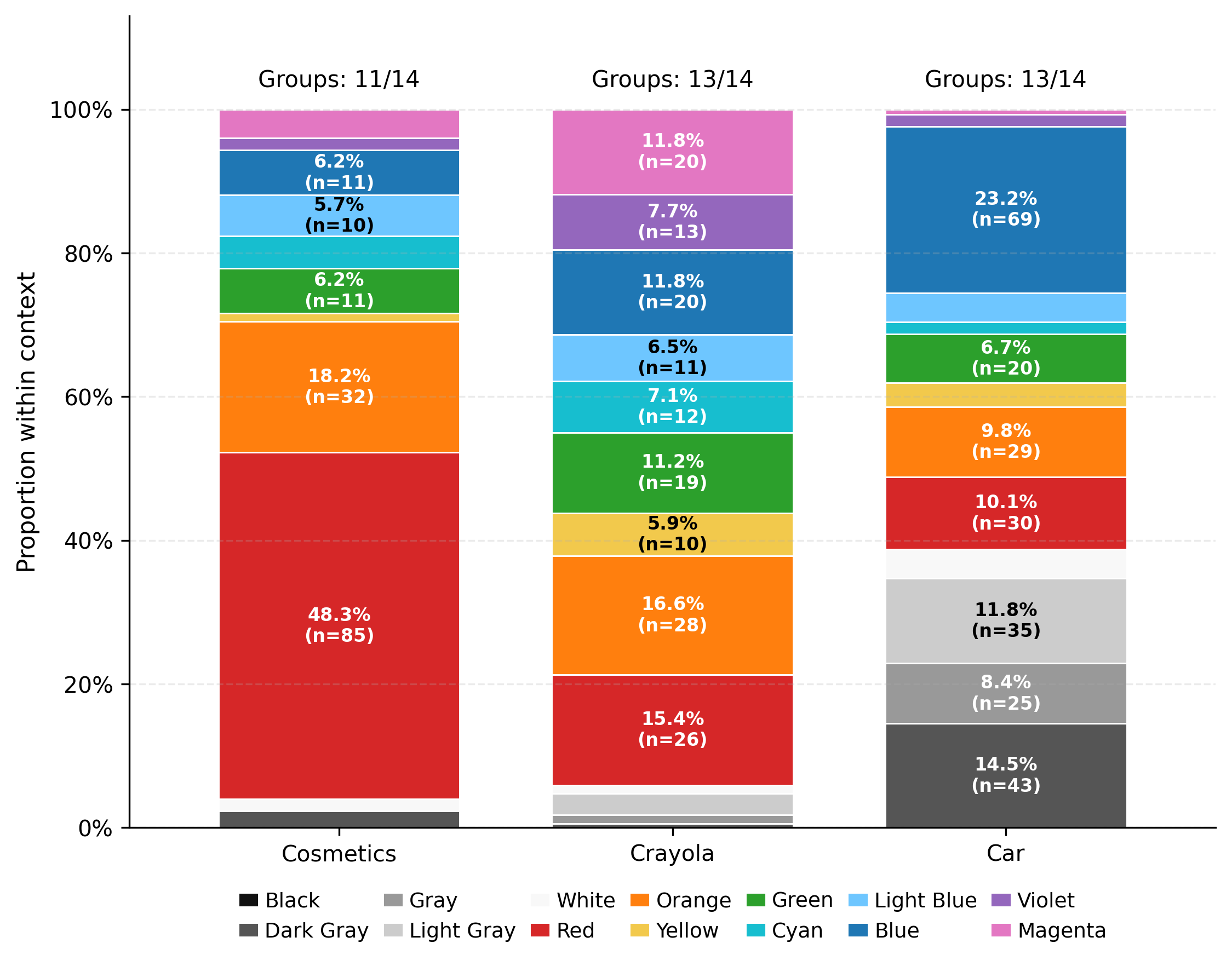}
    \caption{Distribution of Fuzzy Color Groups Across Contexts}
    \label{fig:fuzzy_group_composition}
\end{figure}


While the same COLIBRI color category denotes an identical color region, the number and specificity of associated names vary considerably across application domains. For example, a single color category may correspond to only one name in the car domain, several names in colored pencils, and hundreds of names in cosmetics.

We introduce a context-adaptive naming mechanism while preserving the original COLIBRI color representation. The proposed approach supports two complementary scenarios.

In the first scenario, the representation is simplified by merging the chromatic colors into the nine primary hue groups (red, orange, yellow, green, cyan, light blue, blue, violet, magenta) while preserving the five achromatic colors (black, white, dark gray, gray, and light gray). This produces a compact representation consisting of 14 generalized color categories suitable for applications requiring coarse color discrimination (see Fig.\ref{fig:fuzzy_group_composition}).

\begin{figure}[ht!]
    \centering

    \begin{subfigure}{\linewidth}
        \centering
        \includegraphics[width=\linewidth]{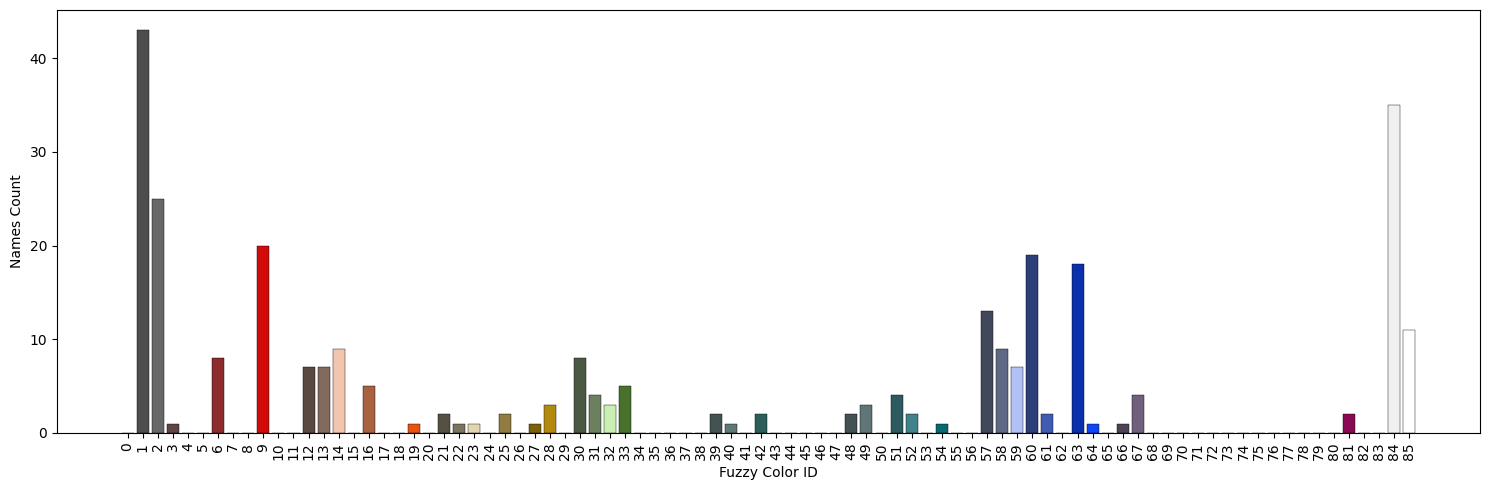}
        \caption{Car colors}
        \label{fig:cars_colors}
    \end{subfigure}

    \vspace{0.5em}

    \begin{subfigure}{\linewidth}
        \centering
        \includegraphics[width=\linewidth]{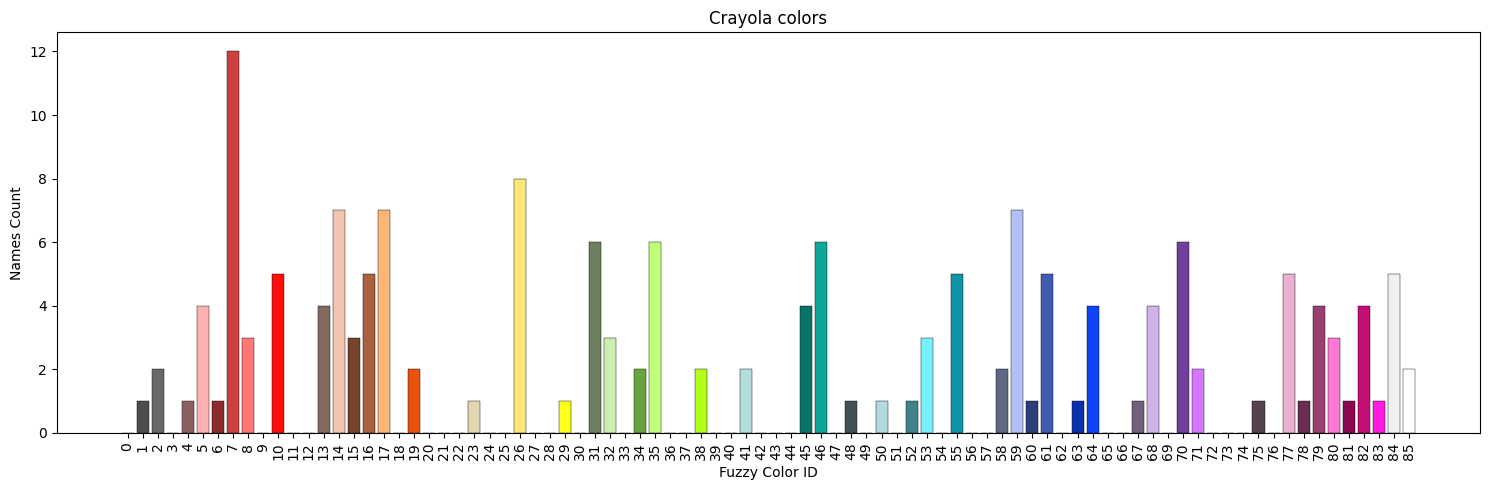}
        \caption{Crayola colors}
        \label{fig:crayola_colors}
    \end{subfigure}

    \vspace{0.5em}

    \begin{subfigure}{\linewidth}
        \centering
        \includegraphics[width=\linewidth]{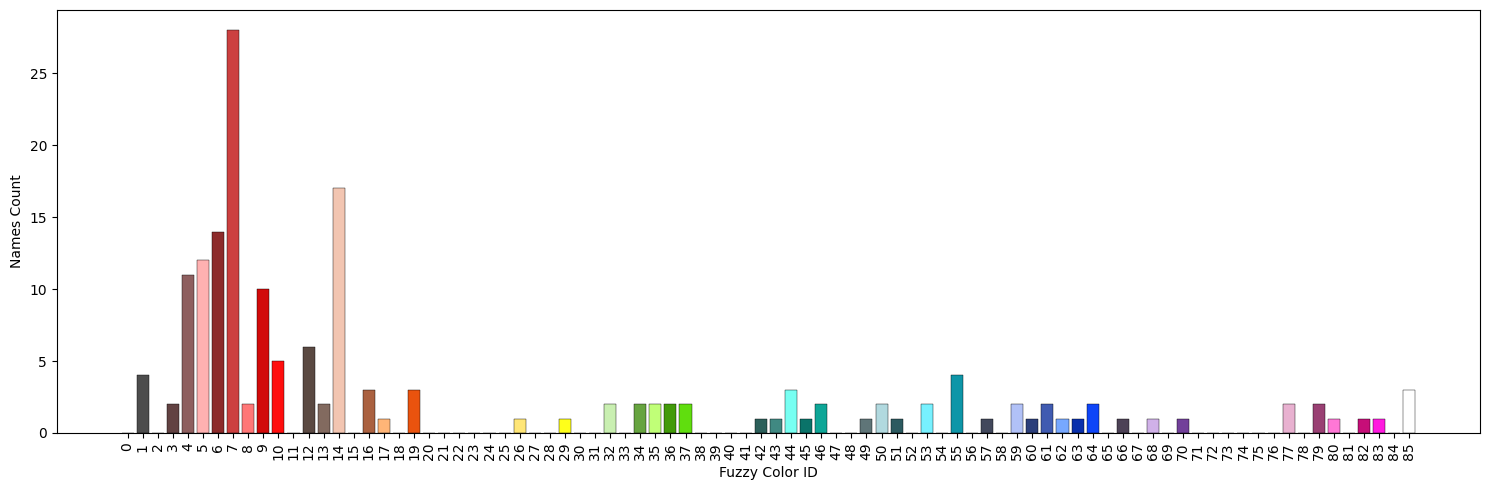}
        \caption{Cosmetics colors}
        \label{fig:html_colors}
    \end{subfigure}

    \caption{Distribution of color names across the 86 COLIBRI color IDs in different semantic contexts. The x-axis represents the COLIBRI color IDs, while the y-axis indicates the number of distinct color names associated with each color.}
    \label{fig:context_distributions}
\end{figure}

In the second scenario, the original 86-color COLIBRI representation is enriched with multiple context-specific names associated with each color category. Instead of introducing additional color classes, the semantic vocabulary is expanded according to the application domain, allowing different naming granularities while maintaining the same underlying color representation (see Fig.\ref{fig:context_distributions}).


\subsubsection{Color Naming Richness Classification}

To interpret the distributional behavior of each semantic context, we define a rule-based color naming richness classification model (see Table~\ref{tab:color_richness_rules}). 

The model uses three complementary indicators: \textit{Coverage, Normalized entropy,} and \textit{Maximum lift}. \textit{Coverage} measures how many fuzzy color regions are represented,\textit{ Normalized entropy} measures how evenly color names are distributed across the represented regions, and \textit{Maximum lift }captures strong overrepresentation of specific fuzzy regions.

Let $K=86$ denote the total number of fuzzy color regions. For each context $d$, let $n_{k,d}$ be the number of color-name records assigned to fuzzy region $k$, and let $N_d$ 
\begin{equation}
N_d = \sum_{k=1}^{K} n_{k,d}
\end{equation}
be the total number of records in context $d$. The \textit{Coverage} ratio is defined as
\begin{equation}
C_d = \frac{|\{k:n_{k,d}>0\}|}{K}.
\label{eq:coverage_ratio}
\end{equation}

The probability of fuzzy region $k$ in context $d$ is defined as
\begin{equation}
p_{k,d} = \frac{n_{k,d}}{N_d}.
\label{eq:region_probability}
\end{equation}

The \textit{Normalized entropy} is computed as
\begin{equation}
H_d^{norm} =
\frac{-\sum_{k:p_{k,d}>0} p_{k,d}\log_2(p_{k,d})}{\log_2(K)}.
\label{eq:normalized_entropy}
\end{equation}

The \textit{Maximum lift }is defined as
\begin{equation}
L_d^{max} =
\max_k \left(\frac{p_{k,d}}{1/K}\right)
=
\max_k \left(Kp_{k,d}\right).
\label{eq:max_lift}
\end{equation}

\begin{table}[!t]
\centering
\caption{Rule-based classification of color naming richness.}
\label{tab:color_richness_rules}
\footnotesize
\renewcommand{\arraystretch}{1.20}
\setlength{\tabcolsep}{2.5pt}
\begin{tabular}{
>{\raggedright\arraybackslash}m{0.24\columnwidth}
>{\raggedright\arraybackslash}m{0.34\columnwidth}
>{\raggedright\arraybackslash}m{0.34\columnwidth}}
\toprule
\textbf{Class} & \textbf{Rule} & \textbf{Interpretation} \\
\midrule
Broad-balanced &
High $C_d$, high $H_d^{norm}$, low/moderate $L_d^{max}$ &
Names are widely and evenly distributed across fuzzy regions. \\

\addlinespace[0.25em]

Broad-concentrated &
High $C_d$, moderate $H_d^{norm}$, high $L_d^{max}$ &
Many regions are covered, but a few regions dominate. \\

\addlinespace[0.25em]

Narrow-specialized &
Low/moderate $C_d$, low/moderate $H_d^{norm}$, high $L_d^{max}$ &
Names focus on a smaller set of domain-specific regions. \\

\addlinespace[0.25em]

Sparse-low richness &
Low $C_d$, low $H_d^{norm}$, low/moderate $L_d^{max}$ &
Few regions are represented and the vocabulary is weakly distributed. \\
\bottomrule
\end{tabular}
\end{table}


\begin{figure*}[ht!]
    \centering
    \includegraphics[width=0.9\textwidth]{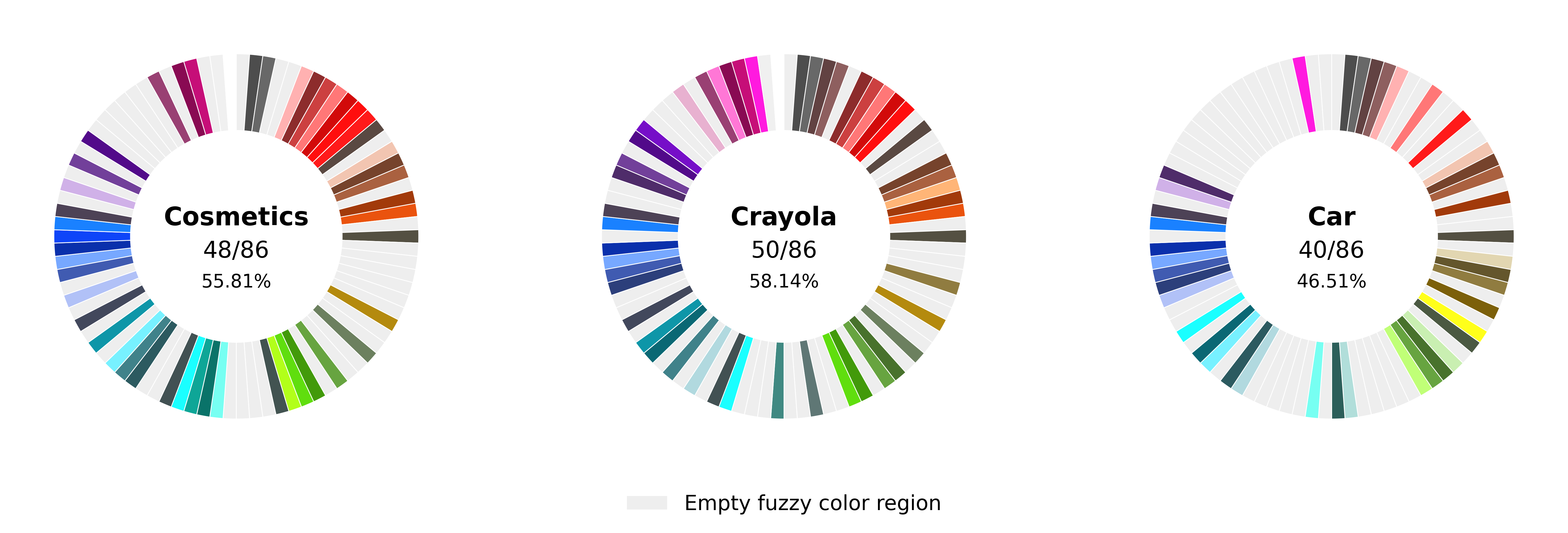}
    \caption{Coverage of 86 Fuzzy Color Regions by Semantic Context}
    \label{fig:fuzzy_clor_coverage}
\end{figure*}

\begin{figure}[ht!]
    \centering
    \includegraphics[width=0.9\columnwidth]{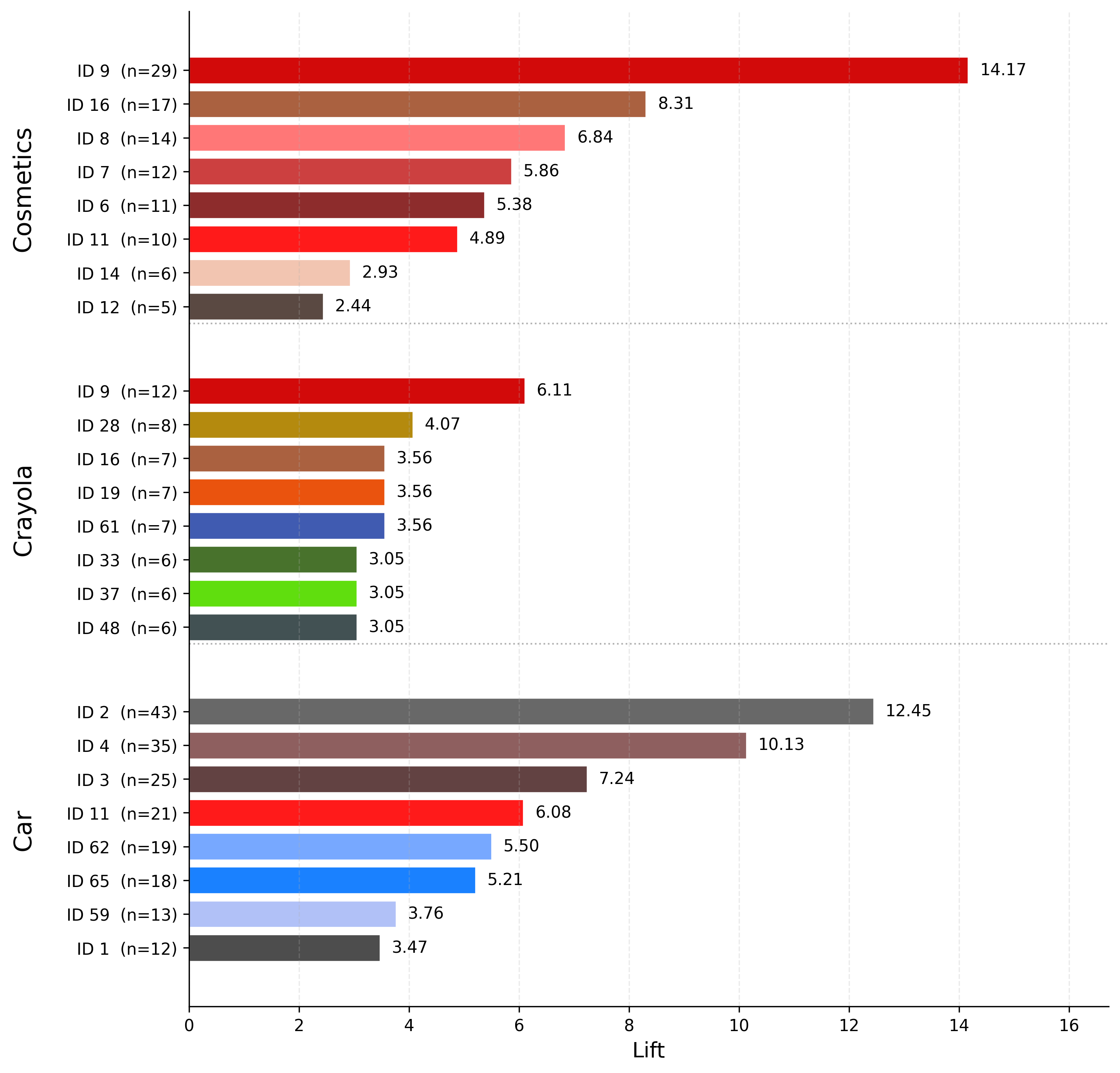}
    \caption{Top Overrepresented Fuzzy Color Regions Across Contexts}
    \label{fig:fuzzy_clor_lift}
\end{figure}

\begin{figure*}[ht!]
    \centering

    \begin{subfigure}[t]{0.46\textwidth}
        \centering
        \includegraphics[width=\textwidth]{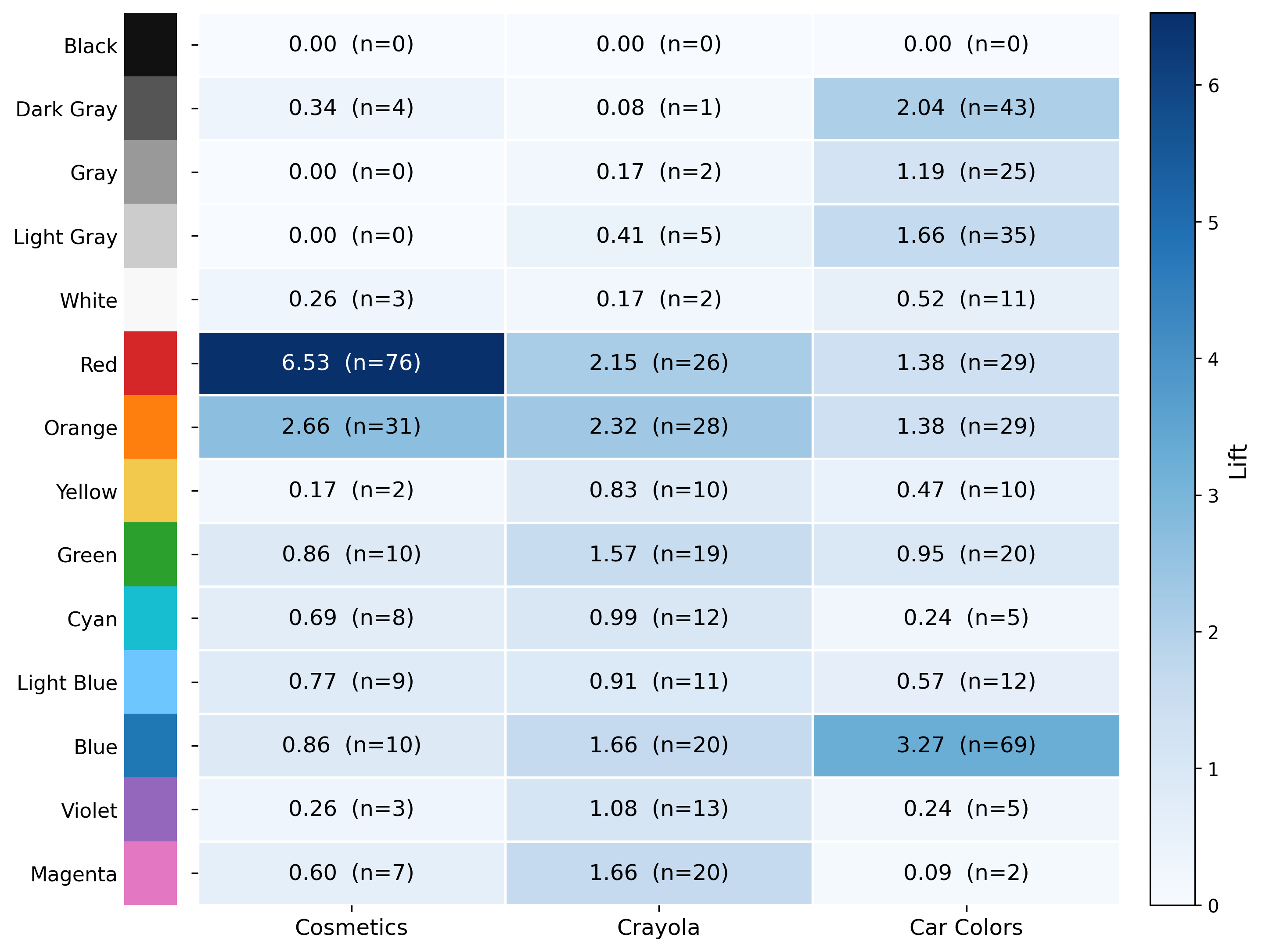}
        \caption{Lift-based overrepresentation of fuzzy color groups.}
        \label{fig:fuzzy_color_lift}
    \end{subfigure}
    \hfill
    \begin{subfigure}[t]{0.46\textwidth}
        \centering
        \includegraphics[width=\textwidth]{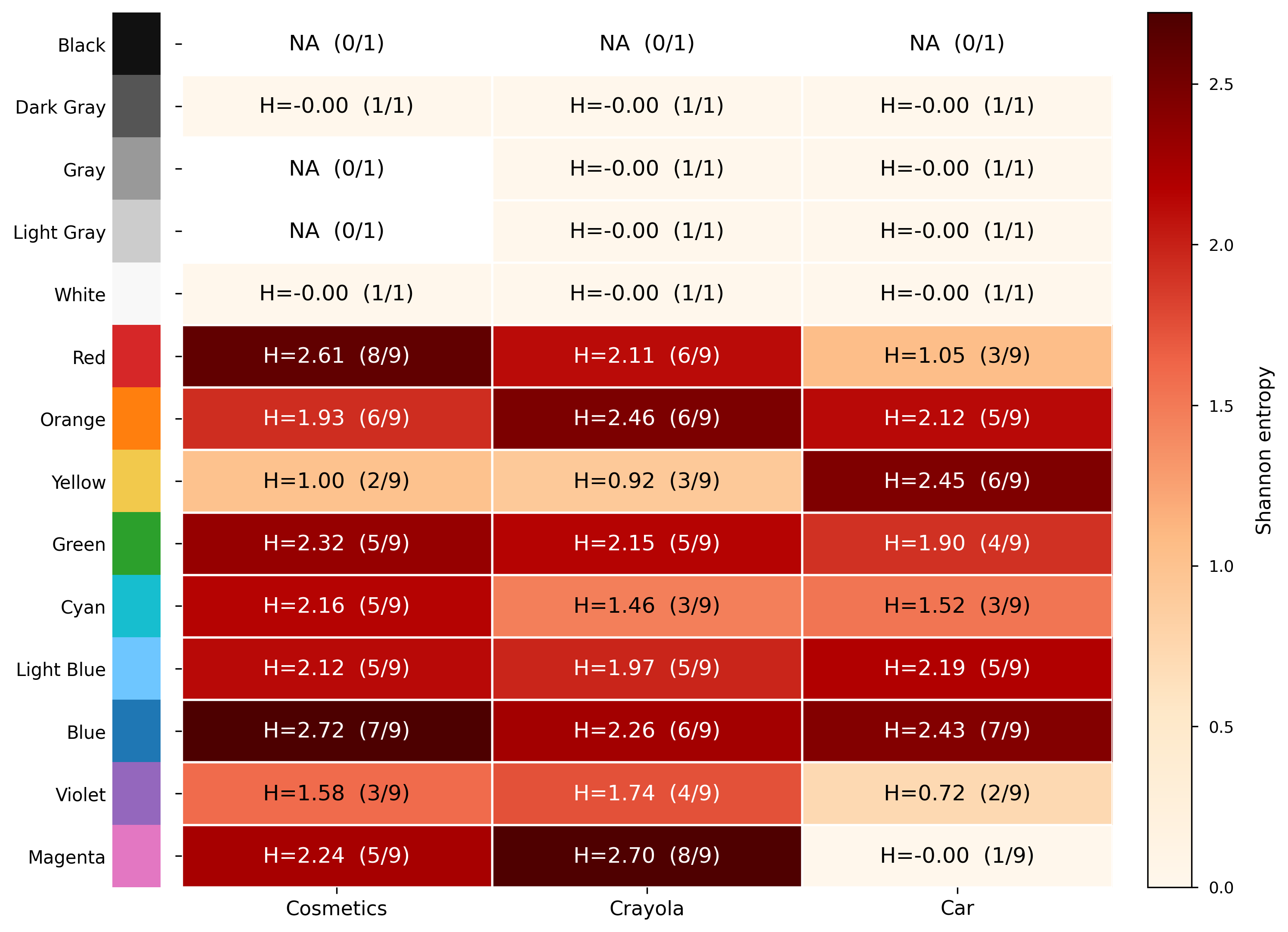}
        \caption{Shannon entropy of fuzzy color groups.}
        \label{fig:fuzzy_color_shannon}
    \end{subfigure}

    \caption{Contextual distribution and internal diversity of fuzzy color groups across Cosmetics, Crayola, and Car Colors.}
    \label{fig:fuzzy_color_lift_entropy}
\end{figure*}

\section{Results}

Fig. \ref{fig:fuzzy_group_composition} and Fig. \ref {fig:context_distributions} compare the distributions of color names across three contexts: \textit{Car}, \textit{Crayola}, and \textit{Cosmetics}. The distributions differ substantially between contexts. The \textit{Crayola} dataset exhibits a similar but sparser distribution, with names concentrated around a subset of color IDs. The \textit{Car }dataset is highly skewed, with a few neutral color IDs (e.g., black, white, blue, and grey regions) accumulating a disproportionately large number of names, while many color IDs have few or no associated names. In contrast, the \textit{Cosmetics} dataset shows the strongest concentration in the red color region, while the others are represented by few names and many remain unused. These results indicate that color naming is strongly context-dependent and that different domains utilize distinct subsets of the COLIBRI color space. 


Table~\ref{tab:coverage_entropy_metrics} provides an overall summary of the fuzzy-color coverage and distributional diversity across the three semantic contexts. Among the analyzed datasets, \textit{Crayola} achieved the broadest coverage of the 86 COLIBRI fuzzy color regions, covering 50 of them (58.14\%). \textit{Cosmetics} covered 48 regions (55.81\%), while \textit{Car} covered only 40 regions (46.51\%). These results are visually confirmed in Fig.~\ref{fig:fuzzy_clor_coverage}. Although the \textit{Car} dataset contains the largest number of records ($n=297$), its color names are distributed over a narrower portion of the fuzzy color space.


The entropy metrics in Table~\ref{tab:coverage_entropy_metrics} show that \textit{Crayola} has the most balanced distribution, with the highest Shannon entropy ($H=5.31$), normalized entropy ($0.83$), and effective number of fuzzy colors ($39.66$). In contrast, \textit{Car} has the lowest entropy ($H=4.48$) and the lowest effective number of fuzzy colors ($22.29$), indicating a stronger concentration in fewer fuzzy regions. \textit{Cosmetics} shows intermediate coverage, but its high maximum probability ($0.16$) and maximum lift ($14.17$) suggest a strong dominance of specific fuzzy color regions.



The overrepresentation-lift analysis provides additional evidence of context specificity. Fig.~\ref{fig:fuzzy_clor_lift} confirms that different contexts emphasize different detailed fuzzy regions. \textit{Cosmetics} is mainly overrepresented by red and orange regions, especially ID~9 with the highest lift value of 14.17. In contrast, \textit{Car} colors are dominated by neutral and blue-related regions, including IDs~2, 4, 3, 62, and 65. This shows that context affects not only overall color coverage, but also which specific fuzzy regions become dominant.

This tendency is also visible at the fuzzy-group level in Fig.~\ref{fig:fuzzy_color_lift}. In \textit{Cosmetics}, red is strongly overrepresented, with a lift value of 6.53, followed by orange (2.66). In \textit{Crayola}, the strongest group-level lifts are observed for orange (2.32) and red (2.15), while blue, green, violet, and magenta remain moderately overrepresented. In \textit{Car}, blue exhibits the strongest group-level lift (3.27), followed by dark gray (2.04) and light gray (1.66). These findings confirm that \textit{Cosmetics} emphasizes warm tones, \textit{Crayola} distributes emphasis across several groups, and \textit{Car} strongly favors blue and achromatic categories (see Fig.~\ref{fig:fuzzy_group_composition}).


Fig.~\ref{fig:fuzzy_color_shannon} highlights the internal diversity of fuzzy color groups. \textit{Cosmetics} shows high diversity mainly in blue ($H=2.72$) and red ($H=2.61$) groups, while \textit{Crayola} is more diverse in magenta ($H=2.70$), orange ($H=2.46$), and blue ($H=2.26$). In \textit{Car} colors, diversity is concentrated in blue ($H=2.43$), yellow ($H=2.45$), and light blue ($H=2.19$) groups, whereas magenta is highly localized. The entropy results show that even within the same broad color group, semantic contexts differ in how widely they use detailed fuzzy subregions.



Next, Table~\ref{tab:coverage_entropy_metrics} summarizes both the quantitative distribution metrics and the resulting richness class for each semantic context.

Overall, the results show that semantic context strongly affects color naming. \textit{Crayola} uses the fuzzy color space more broadly and evenly, \textit{Cosmetics} is concentrated around warm-tone regions, and \textit{Car} colors are more specialized around blue and achromatic tones. These differences suggest that context-specific color vocabularies are more appropriate than a single universal set of color names.

\section{Conclusion}

This study shows that color naming is strongly influenced by semantic context. Using the 86 COLIBRI fuzzy color representation, the results demonstrated that \textit{Crayola} provides the broadest and most balanced use of the fuzzy color space, \textit{Cosmetics} is mainly concentrated around red and orange regions, and \textit{Car} colors are more specialized around blue and achromatic regions. Consumer research shows that unusual, evocative names improve product evaluation more than generic ones \cite{Chou2020}, explaining why \textit{Cosmetics} multiplies names in its dominant hues while \textit{Car}, a more utilitarian choice, keeps labels compact -- consistent with prior computational work showing that universal vocabularies fail to capture how color is used across domains \cite{Yu_Lu_2018}.

The coverage, lift, and entropy analyses confirm that the same fuzzy color representation can capture different naming structures across domains. These findings suggest that context-specific color vocabularies are more appropriate than relying on a single universal set of color names.

This study has several limitations. Sample sizes differ across domains (176, 169, and 297), and since coverage is more sensitive to sample size than entropy, part of \textit{Car}'s lower coverage may reflect its smaller name pool rather than domain-specific concentration alone. Only three domains were analyzed, limiting the extent to which the coverage-entropy relationship can be generalized.

\section*{Acknowledgment}
This research has been funded by the Science Committee of the Ministry of Science and Higher Education of the Republic of Kazakhstan (Grant No. AP22786412)

\bibliographystyle{IEEEtran}
\bibliography{bibfile}

\end{document}